\documentclass[authoryear]{article}

\usepackage[authoryear]{natbib}
\usepackage{arxiv}
\usepackage[utf8x]{inputenc}
\usepackage[T1]{fontenc}
\usepackage{mathtools}
\usepackage{amsmath}

\usepackage{amssymb}
\usepackage{graphicx}
\usepackage{titling}
\usepackage{ulem}
\usepackage{afterpage}

\title{Recombination of Artificial Neural Networks}
\author{Aaron Vose, Jacob Balma, Alex Heye, Alessandro Rigazzi, Charles Siegel, Diana Moise, \\Benjamin Robbins, and Rangan Sukumar \\Cray Inc., 901 Fifth Avenue, Suite 1000, Seattle, WA 98164, USA}
\begin{document}
\maketitle

\begin{abstract}
We propose a genetic algorithm (GA) for hyperparameter optimization of
artificial \linebreak neural networks which includes chromosomal
crossover as well as a decoupling \linebreak of parameters (i.e.,
weights and biases) from hyperparameters (e.g., learning rate,
\linebreak weight decay, and dropout) during sexual reproduction.
Children are produced \linebreak from three parents; two contributing
hyperparameters and one contributing the \linebreak parameters. Our
version of population-based training (PBT) combines traditional
\linebreak gradient-based approaches such as stochastic gradient
descent (SGD) with our \linebreak GA to optimize both parameters and
hyperparameters across SGD epochs.  Our \linebreak improvements over
traditional PBT provide an increased speed of adaptation and
\linebreak a greater ability to shed deleterious genes from the
population. Our methods \linebreak improve final accuracy as well
as time to fixed accuracy on a wide range of deep neural network
architectures including convolutional neural networks, recurrent
neural networks, dense neural networks, and capsule networks.
\end{abstract}

\vspace{2.0\baselineskip}

\section{Introduction}
As deep learning practitioners craft artificial neural networks (NNs) to solve real-world problems, some parts of the NN architectures may be fairly clear, such as the desired inputs and outputs, but other details, such as the specifics of the intermediate layers, are unknown with no clear method available with which to derive them.
While the weights and biases of a NN, referred to as model parameters, are commonly computed by derivative-based approaches such as stochastic gradient descent (SGD), finding the optimal topology and training regime, referred to as hyperparameters (HPs), is an unsolved problem and an area of active research. Tuning these neural network HPs can result in a higher final accuracy and$\,$/$\,$or faster training to a fixed accuracy. In extreme cases, training may not converge without a good choice of hyperparameters.

\begin{figure}[!h!]
\centering
\includegraphics[width=0.75\textwidth]{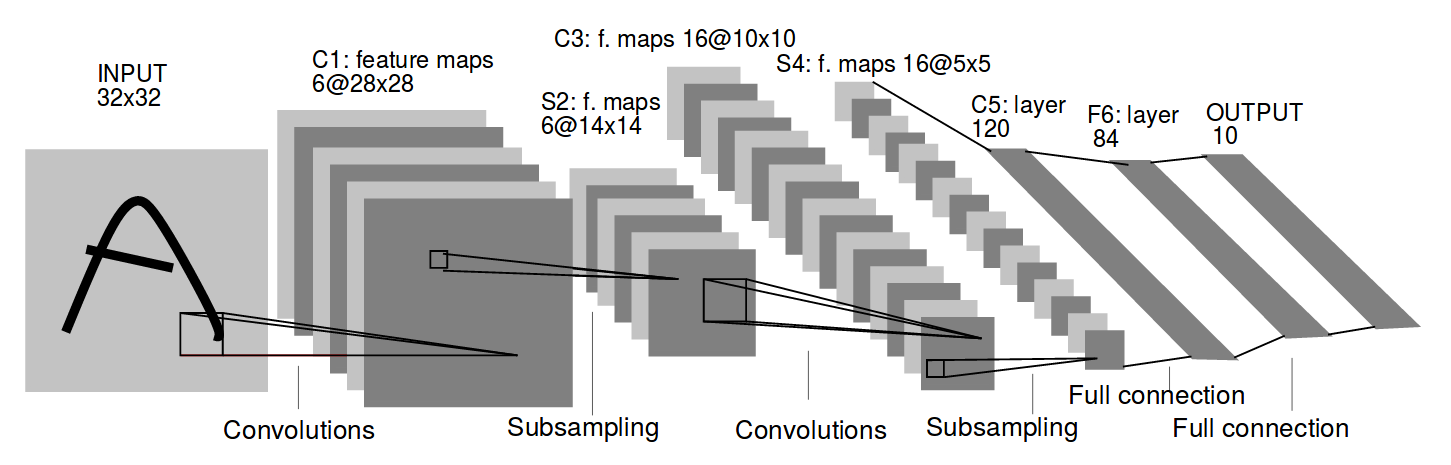}
\caption{\label{fig:lenet}The LeNet convolutional neural network architecture is shown, representing a relatively small neural network with many thousands of possible hyperparameter sets (image from \cite{lecun1998gradient}).}
\end{figure}

The search space of hyperparameter sets can be enormous even for small convolutional networks such as those using the LeNet architecture shown in Figure \ref{fig:lenet} \cite{lecun1998gradient}.  Indeed, naive approaches like random and grid search are often computationally intractable due to the number of trials required to find a good HP set.
For larger NNs, there can easily be more hyperparameter sets than atoms in the observable universe, motivating the use of optimization strategies which can prune undesirable regions of the search space. 
This work introduces a biologically-inspired genetic algorithm (GA) customized for hyperparameter optimization. While GA details are covered in Section \ref{sect:HPO_GA}, one might think of a GA applied to HPO as being ``iterative, parallel, stochastic grid-search with pruning.''

Section \ref{sect:experiments} demonstrates the effectiveness of our genetic algorithm for the optimization of example NNs with two different approaches: Section \ref{sect:exp_traditional} covers ``traditional'' HPO including topology HPs, and Section \ref{sect:hpo_pbt} covers population-based training (PBT) for the creation of dynamic training schedules \cite{pbt2017}.
We consider three case studies for each of the two approaches and provide results showing improvements in final accuracy as well as time to fixed accuracy for a wide range of deep neural network architectures including convolutional neural networks (CNNs), recurrent neural networks (RNNs), dense neural networks, and capsule networks.

\section{Related Work}
Previous work addressing hyperparameter optimization in a high-performance computing setting spans a wide range of complexities and efficacies, from simple approaches like random search and grid search to more complicated approaches such as \textit{Hyperband} \cite{li2016hyperband} and \textit{Livermore Tournament Fast Batch Learning} (LTFB) \cite{jacobs2017towards}.
While some approaches, such as \textit{Multi-node Evolutionary Neural Networks for Deep Learning} (MENNDL) \cite{potok2016study,young2015optimizing}, explicitly mention the use of genetic algorithms, others, such as \textit{population-based training} (PBT) \cite{pbt2017}, are not expressly called out as such in the literature, even though they appear to be members of a generalized class of genetic algorithm.

Much of this related work appears to place less emphasis on the quality of the optimization algorithm than in implementation details. For example, MENNDL is described by its authors as using a very simple GA, while PBT lacks features such as sexual reproduction with crossover. Such HPO approaches can generally run at large node counts, but a lack of an efficient search algorithm can mean that most of those compute cycles are wasted on poorly-performing areas of the hyperparameter space. 

\begin{figure}[!ht]
\centering
\includegraphics[width=0.575\textwidth]{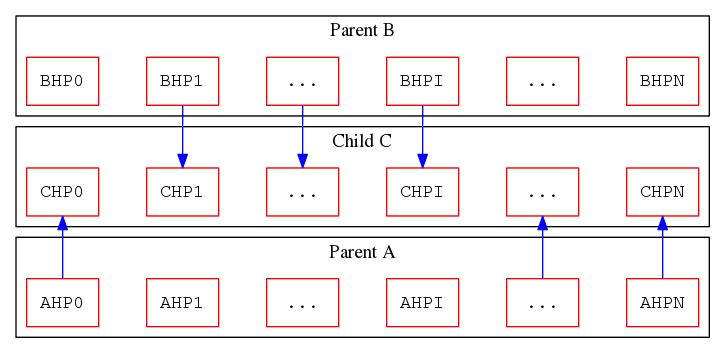}
\caption{\label{fig:crossover}Crossover combines hyperparameters from two parents to create those of a child.} 
\end{figure}

This work first frames PBT as a genetic algorithm and then generalizes and extends it in Section \ref{sect:HPO_GA}. The original version of PBT can be thought of as a genetic algorithm with the following characteristics: overlapping generations, genes comprised of hyperparameters and parameters, GA generations running concurrently with SGD training epochs, asexual reproduction without crossover, two possibilities for viability selection (tournament and rank), and a fitness function customized for each of the NN objectives (e.g., using the BLUE score for neural machine translation \cite{papineni2002bleu}).
Crossover (i.e., recombination), discussed in more detail in Section \ref{sect:HPO_GA}, combines hyperparameters from two parents to create a new child, as depicted in Figure \ref{fig:crossover}. This provides not only a possibility for the aggregation of beneficial genes, but helps shed deleterious mutations from the population.

\section{Genetic Algorithm Details} \label{sect:HPO_GA}

\subsection{Algorithm Details} \label{sect:HPO_GA_AD}
Our GA is motivated by previous work in ecology and theoretical biology with a focus on population genetics \cite{gavrilets2005dynamic,gavrilets2009dynamic,birand2011patterns}. In particular, work modeling the transmission of memes alongside genes provides impetus for a decoupling of the source of NN parameters from the source(s) of the NN hyperparameters \cite{gavrilets2006dynamics}.
This work regards memes as a second type of gene; the first type concerns hyperparameters whereas the second type concerns parameters. We refer to the sources of genetic information for a child as the three parents of that child.

Our genetic algorithm generalizes and extends PBT, incorporating sexual reproduction with crossover as well as a decoupling (``tri-parent'') of the source of hyperparameters and parameters. Chromosomal crossover, depicted in Figure \ref{fig:crossover}, exchanges genetic material between homologous chromosomes from parents during sexual reproduction, resulting in recombinant chromosomes in the offspring.
This provides two primary advantages over asexual reproduction without crossover: an increased speed of adaptation and an increased ability to shed deleterious genes from the population \cite{crow1994advantages}. In the context of hyperparameter optimization, crossover is expected to become increasingly helpful as the number of hyperparameters grows.

\begin{figure}[!ht]
\centering
\fbox{%
\begin{minipage}{4 in}
\vspace{-0.5\baselineskip}
\begin{small}
\null $g \gets 0$ \\
\null $\mathbb{P}_{g} \gets$ initial\_population \\
\null while $g < $ {\small PARAM\_GENERATIONS:} \\
\null \quad for each $p$ in $\mathbb{P}_{g}$: \\
\null \quad \quad $p.\mathrm{\textit{fom}} \gets $ execute$\!\:$( $p$ )\\ 
\null \quad \quad $p.\mathrm{\textit{fitness}} \gets e^{-\sigma((p.\mathrm{\textit{fom}}-\mathrm{\textit{minfom}})/(\mathrm{\textit{maxfom}}-\mathrm{\textit{minfom}}))^2}$\\
\null \quad $\mathbb{P}_{(g+1)}$ $\gets$ $\emptyset$\\
\null \quad while $\left|\mathbb{P}_{(g+1)}\right|$ $<$ {\small PARAM\_POPULATION\_SIZE:} \\
\null \quad \quad $a$ $\gets$ choose $p$ from $\mathbb{P}_{g}$ with probability proportional to $p.\mathrm{\textit{fitness}}$ \\
\null \quad \quad $b$ $\gets$ choose $p$ from $\mathbb{P}_{g}$ with probability proportional to $p.\mathrm{\textit{fitness}}$ \\
\null \quad \quad $\alpha$ $\gets$ choose $p$ from $\mathbb{P}_{g}$ with probability proportional to $p.\mathrm{\textit{fitness}}$ \\
\null \quad \quad $c.\mathrm{\textit{hparams}}$ $\gets$ mutate$\!\:$( crossover$\!\;$( $a.\mathrm{\textit{hparams}}, \: b.\mathrm{\textit{hparams}}$ ) ) \\
\null \quad \quad $c.\mathrm{\textit{params}}$ $\gets$ $\alpha.\mathrm{\textit{params}}$ \\
\null \quad \quad $\mathbb{P}_{(g+1)} \gets \mathbb{P}_{(g+1)} \cup \{c\}$ \\
\null \quad $g \gets g+1$ \\
\end{small}
\vspace{-0.75\baselineskip}
\end{minipage}}
\caption{\label{fig:ga}Genetic algorithm details; when running without PBT, one can ignore $\alpha$.}
\end{figure}

Salient details of the GA presented in this work are shown in Figure \ref{fig:ga}. The most important aspects detailed therein are the use of: probabilistic viability$\,$/$\,$sexual selection, multi-point crossover, and three parents per child.
To be clear, \textit{minfom} and \textit{maxfom} in Figure \ref{fig:ga} refer to the minimum and maximum of the current generation's population members' figures of merit (\textit{fom}), respectively. The \textit{fom} to \textit{fitness} mapping results in population members having a \textit{relative} fitness score, which we find helps to remove problems associated with the fine-tuning of fitness$\,$/$\,$objective functions.
When a locus mutates, its value changes in one of four possible ways with equal probability: decrease by 0--1\%, increase by 0--1\%, increase by 10--20\%, or decrease by 10--20\%.

\subsection{Implementation Details} \label{sect:HPO_GA_ID}
The core genetic algorithm described in Section \ref{sect:HPO_GA_AD} is implemented as a library written in Python \cite{Rossum1995python}. This language choice is popular with many data scientists and also appropriate from a computational point of view; our implementation wraps existing heavily-optimized frameworks such as TensorFlow \cite{tensorflow2015-whitepaper} for the time-consuming SGD work, while the relatively light-weight GA can run as Python code without becoming a bottleneck.
Indeed, the only parts of the computation which really need to be distributed are the SGD training and fitness evaluation. Thus, our implementation effectively utilizes distributed Dask \cite{dask2015_matthew_rocklin-proc-scipy} to run training and evaluation of GA population members in parallel. For maximum scalability, population-level parallelism in the GA can be combined with data-parallel training within each individual NN training instance.
With a proven-effective population size of 100 individuals and data-parallel training of each NN at a scale of 100 nodes, our HPO solution is expected to scale easily to 10,000 compute nodes.

The HPO experiments presented in the following sections were performed with identical settings controlling the behavior of the genetic algorithm. Specifically, a population size of 100 individuals (i.e., neural networks including parameters and hyperparameters), a mutation rate of 0.05 per locus, a crossover rate of 0.33 per locus, and one GA generation per NN training epoch when running with PBT. When not using PBT, the GA is run either for 250 generations or until a 24-hour time limit is reached.
While these settings may be optimized for specific use cases, this work shows that our GA is effective across NNs and datasets without fine-tuning.

\section{Experiments} \label{sect:experiments}

\subsection{Experimental Platforms} \label{sect:exp_platform}
As most experiments reported in this paper deal with the optimization of converged final accuracy, the platform used for NN training is not of paramount importance. This works shows effective HPO across a range of systems with diverse hardware configurations.
Runs were performed on \textit{Cicero} (an in-house Cray XC40 with dual-CPU Ivybridge nodes), \textit{Tiger} (an in-house Cray XC50 with K40 GPU nodes and P100 GPU nodes), as well as \textit{Cori} (NERSC's XC40 with dual-CPU Haswell nodes and KNL accelerator nodes).

\subsection{HPO with Topology} \label{sect:exp_traditional}
This section deals with a traditional approach using simultaneous optimization of hyperparameters controlling NN topology specifics such as the number of layers, each layer's type, and the number of neurons in each layer, as well as training HPs such as learning rate and dropout. In this context, a traditional approach generally trains a neural network to convergence or to a fixed accuracy without a change in the HPs being optimized. In particular, a fixed decay of the learning rate across SGD epochs represents holding the initial learning rate and the decay rate constant.
The principal advantage of this approach as compared to PBT (covered in more detail in Section \ref{sect:hpo_pbt}) is that it can optimize topology. The downside is that it cannot easily create a dynamic HP schedule over NN training epochs.
Thus, we suggest a two-phase HPO approach: a reasonable topology and set of training HPs is first discovered with traditional HPO, and then PBT is used to fine-tune the training schedule while simultaneously optimizing model parameters.

\subsubsection{CANDLE} \label{sect:candle}
\begin{figure}[!ht]
\centering
\includegraphics[width=0.35\textwidth]{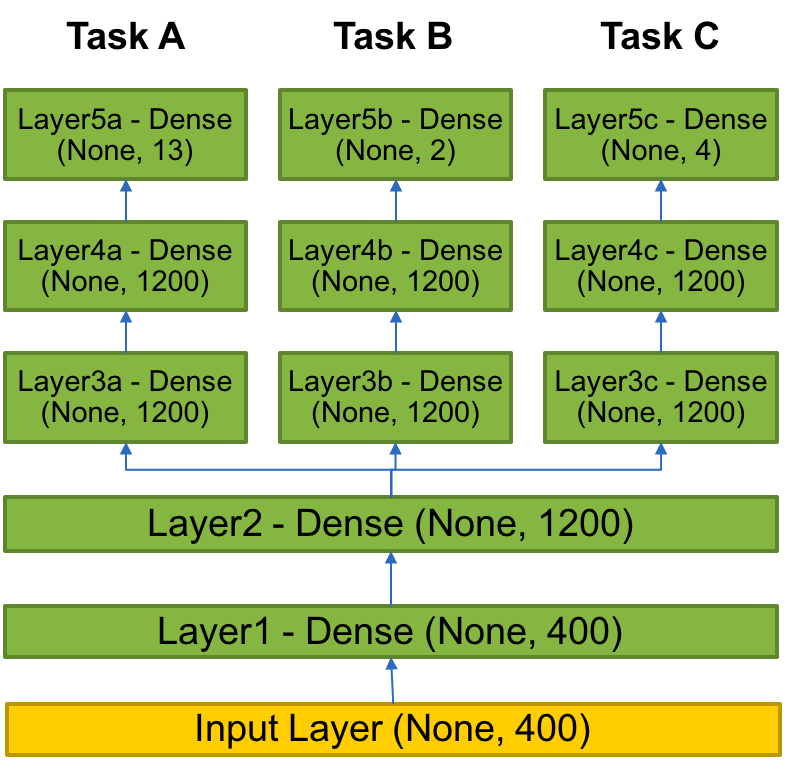}
\caption{\label{fig:candle_p3b1_arch}Architecture of the neural network in CANDLE benchmark P3B1 (image from \cite{ecp_candle_bmks}).}
\end{figure}

The \textit{CANcer Distributed Learning Environment} (CANDLE) benchmark P3B1 uses a multi-task deep neural network (DNN) architecture for a data extraction task applied to clinical reports \cite{ecp_candle_bmks,ecp_candle_article}.
As opposed to training separate NN models for individual machine learning tasks, P3B1 is designed to exploit task-relatedness to simultaneously learn multiple concepts. As seen in Figure \ref{fig:candle_p3b1_arch}, the NN architecture relies heavily on stacks of fully-connected layers.
The data set is a sample of around 1,000 annotated pathology reports, each with an input dimension of about 200--500 as a bag of concepts, created by preprocessing original reports with a bag of characters size of 250,000--500,000 and a bag of words size of 5,000--20,000. The three outputs represent the three tasks, dealing with tumor sites, tumor laterality, and clinical grade of tumors.

In this experiment, HPO is performed with P3B1 on \textit{Tiger} (see Section \ref{sect:exp_platform}) to minimize the training time required to hit a fixed accuracy. Specifically, a minimum F1 macro and micro score must be met for each of the three tasks: primary site (0.275, 0.506), tumor laterality (0.907, 0.907), and histological grade (0.485, 0.647).
Using this figure of merit tends to help discover HP sets with good training HPs as well as smaller NN topologies which are able to reach the same or higher accuracies as larger models. This also indirectly reduces the computational cost of using the trained NN for inference.
Table \ref{tab:candle_hps} shows the hyperparameters optimized in this experiment along with their optimized values. The layer names used in the table correspond to the layers identified in Figure \ref{fig:candle_p3b1_arch}.

\begin{table}[ht!]
\centering
\begin{small}
\begin{tabular}{l|rrr|r}
\textbf{Hyperparameter} & \textbf{Min} & \textbf{Max} & \textbf{Reported in \cite{ecp_candle_bmks}} & \textbf{Optimized} \\
\hline
$\vphantom{|^|}$\textit{dense\_sz$_{A3,A4}$}          & 32 & 8192 & 1200 & 324 \\
\textit{dense\_sz$_{B3,B4}$}          & 32 & 8192 & 1200 & 279 \\
\textit{dense\_sz$_{C3,C4}$}          & 32 & 8192 & 1200 & 359 \\
\textit{dense\_sz$_2$}          & 32 & 8192 & 1200 &  459 \\
\hline
$\vphantom{|^{|^|}}$\textit{learning\_rate}         & 1.0$e^{-8}$ & 1.0$e^{-0}$ & 1.0$e^{-2}$ & 2.1$e^{-2}$ \\
\textit{dropout}                  & 0.0$e^{-0}$ & 9.0$e^{-1}$ & 1.0$e^{-1}$ & 8.8$e^{-2}$ \\
\end{tabular}
\end{small}
\vspace{.25\baselineskip}
\caption{\label{tab:candle_hps}Hyperparameters used for HPO on the CANDLE P3B1 neural network.}
\end{table}

The optimized hyperparameters result in a neural network that is able to be trained to the target accuracies in 25.8 seconds, while the original network takes 118.6 seconds (average of 10 training runs).
These results show that a smaller network is able to be trained in 78.2\% less time than the original. The discovered hyperparameter set has a learning rate which is double the initial, whereas the change in dropout is minor and does not significantly impact NN training performance. Indeed, most of the reduction in training time comes from the changes to topology.
When optimizing with training time to fixed accuracy as the figure of merit, it is important to set target accuracies that are high enough for the NN's production use case, as a smaller network may be able to reach lower accuracies faster than a larger network, but not be able to reach as high of a final accuracy.
Fixed accuracy figure of merit is particularly interesting for the creation of smaller neural networks for inferencing applications where computational performance is limited, such as in the embedded and mobile markets.

\subsubsection{RPV on Pythia+Delphes ATLAS Data} \label{sect:PRV_atlas}
\begin{figure}[!ht]
\centering
\includegraphics[width=0.85\textwidth]{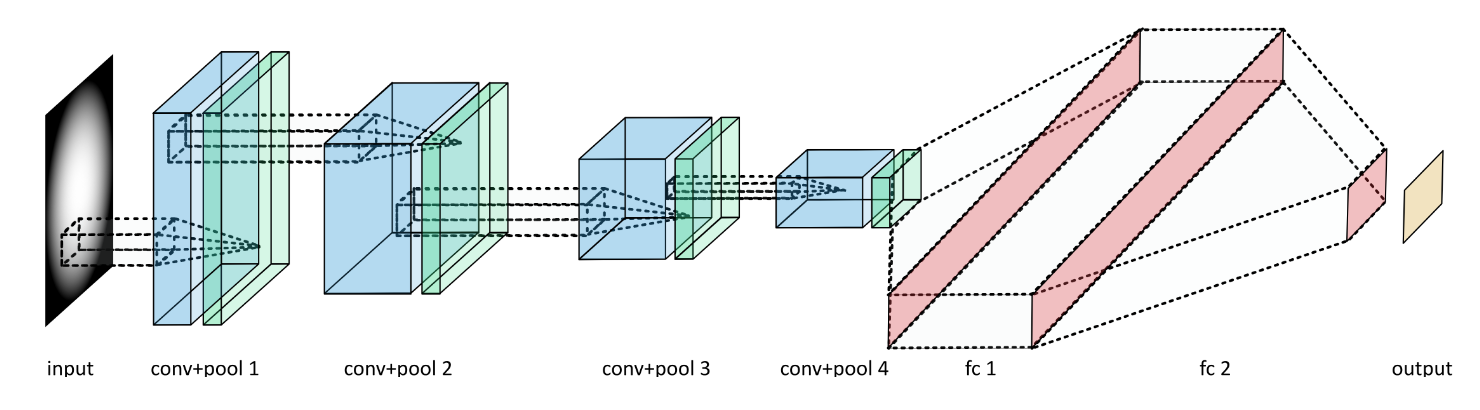}
\caption{\label{fig:rpv_arch}High-level architecture of the RPV convolutional neural network (image from \cite{bhimji2017deep}).}
\end{figure}

The use-case of the RPV neural network explored in this section deals with searching for massive supersymmetric (`RPV-Susy') particles in multi-jet final states at the Large Hadron Collider \cite{bhimji2017deep}.
The data set in this example is produced by the Pythia \cite{sjostrand2008brief} event generator interfaced to the Delphes \cite{de2014delphes} fast detector simulation. The NN classifies events as either belonging to the RPV-Susy signal (i.e., new-physics) or belonging to the most prevalent background (i.e., `QCD').
The RPV architecture depicted in Figure \ref{fig:rpv_arch} is a convolutional neural network, trained on $64\times64$-pixel sparse images with three channels. This work uses a data set with 64,000 training items and 32,000 test items during HPO, while the full training set of 412,416 and full test set of 137,471 items are used for final evaluation after HPO.

In this experiment, HPO is performed on \textit{Cori} (see Section \ref{sect:exp_platform}) with the hyperparameters presented in Table \ref{tab:rpv_hps} as well as the optimizer hyperparameter ($\textit{optimizer} \in \{$Adam, Nadam, Adadelta$\}$). Each NN is trained for 16 epochs during evaluation, which is long enough to nearly reach convergence.
Our GA discovered a new topology and produced training HPs able to achieve better final accuracy (\textit{optimizer} did not change from the original Adam).  Evaluation using the full test set of 137,471 examples shows the original HPs achieve 9.22\% error, whereas our optimized HPs attain 1.78\% error. These results represent an average from three independent training runs and are stable to within less than a half of a percent.

\begin{table}[ht!]
\centering
\begin{small}
\begin{tabular}{l|rrr|r}
\textbf{Hyperparameter} & \textbf{Min} & \textbf{Max} & \textbf{Reported in \cite{bhimji2017deep}} & \textbf{Optimized} \\
\hline
$\vphantom{|^|}$\textit{conv\_kernel\_sz$_1$} & 2  & 9   & 3   & 4  \\
\textit{conv\_kernel\_sz$_2$} & 2  & 9   & 3   & 4  \\
\textit{conv\_kernel\_sz$_3$} & 2  & 9   & 3   & 4  \\
\textit{conv\_filters$_1$}      & 2  & 32  & 8   & 8  \\
\textit{conv\_filters$_2$}      & 4  & 64  & 16  & 15 \\
\textit{conv\_filters$_3$}      & 8  & 128 & 32  & 28 \\
\textit{dense\_sz}              & 16 & 256 & 64  & 59 \\
\hline
$\vphantom{|^{|^|}}$\textit{learning\_rate}         & 1.0$e^{-5}$ & 1.0$e^{-1}$ & 1.0$e^{-3}$ & 1.0$e^{-1}$ \\
\textit{dropout}                  & 1.0$e^{-3}$ & 9.0$e^{-1}$ & 5.0$e^{-1}$ & 5.0$e^{-1}$ \\
\end{tabular}
\end{small}
\vspace{.25\baselineskip}
\caption{\label{tab:rpv_hps}Hyperparameters used for HPO on the RPV convolutional neural network.}
\end{table}

\subsubsection{LeNet on MNIST} \label{sect:lenet_mnist}
\begin{figure}[!ht]
\centering
\includegraphics[width=0.75\textwidth]{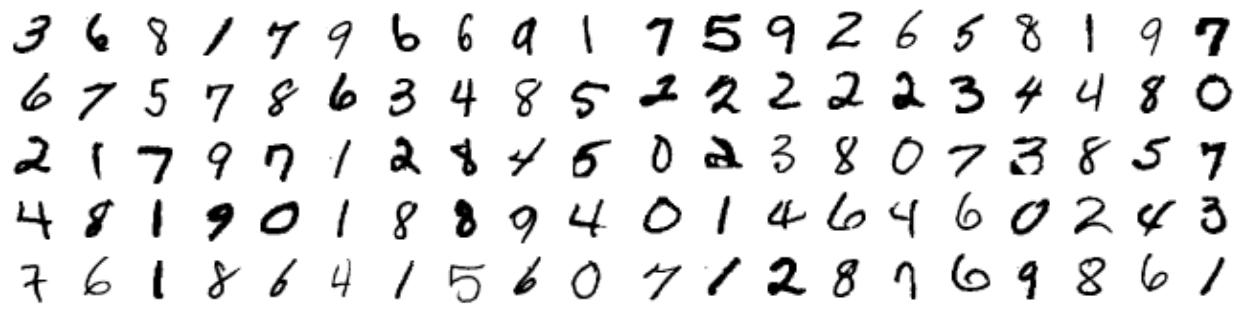}
\caption{\label{fig:mnist}Selected examples from the MNIST dataset of handwritten digits (images from \cite{lecun1998mnist}).}
\end{figure}

The LeNet architecture, visualized in Figure \ref{fig:lenet}, is a simple neural network introduced by Yann LeCun to demonstrate the power of convolutional neural networks (CNNs) applied to the task of image classification \cite{lecun1998mnist}. It consists of two convolutional layers, each of which is followed by a subsampling layer, and then a pair of fully-connected layers with a final output layer.
LeNet was designed to classify the handwritten digit images in the MNIST dataset, samples of which are provided in Figure \ref{fig:mnist}. This dataset contains 70,000 labeled images of handwritten numerals; each single channel, $28 \times 28$ pixel resolution. Ten thousand images are held out to form a testing set. 

A version of this architecture is used to evaluate training time to a fixed accuracy on \textit{Cicero} (see Section \ref{sect:exp_platform}) using our GA for HPO. This experiment targets the topology and training hyperparameters listed in Table \ref{tab:lenet_hps} and uses a validation accuracy threshold of 98.6\%, motivated by HPO experiments in previous work \cite{ahaye2018_scaledl}. To remove noise when timing training of the NN, the NN is not considered to have passed the threshold until the target validation accuracy is met on two successive validation evaluations (performed once every 100 mini-batches).
This HPO run discovered a new, smaller NN topology which combined with optimized training hyperparameters can reach the target accuracy in 83.5\% less time than the original HPs. The new hyperparameter set takes an average of 32.4 seconds to train, while the original HP set requires 164.5 seconds on average across ten runs.

\begin{table}[ht!]
\centering
\begin{small}
\begin{tabular}{l|rrr|r}
\textbf{Hyperparameter} & \textbf{Min} & \textbf{Max} & \textbf{Reported in \cite{ahaye2018_scaledl}} & \textbf{Optimized} \\
\hline
$\vphantom{|^|}$\textit{conv\_kernel\_sz$_1$} & 2  & 8   & 5   & 4   \\
      \textit{conv\_kernel\_sz$_2$} & 2  & 8   & 5   & 4   \\
      \textit{conv\_filters$_1$}      & 8  & 128  & 32   & 8  \\
      \textit{conv\_filters$_2$}      & 16  & 256  & 64  & 18  \\
      \textit{dense\_sz}              & 64  & 4096 & 1024 & 1024 \\
      \hline
      $\vphantom{|^{|^|}}$\textit{learning\_rate}         & 1.0$e^{-6}$ & 1.0$e^{-2}$ & 1.0$e^{-4}$ & 4.3$e^{-4}$ \\
      \textit{dropout}                  & 5.0$e^{-3}$ & 9.0$e^{-1}$ & 5.0$e^{-1}$ & 8.3$e^{-1}$ \\
\end{tabular}
\end{small}
\vspace{.25\baselineskip}
\caption{\label{tab:lenet_hps}Hyperparameters used for HPO on LeNet with the MNIST dataset.}
\end{table}

\subsection{HPO with PBT} \label{sect:hpo_pbt}
Hyperparameter optimization with population-based training has the advantage of not only creating a HP schedule over NN training epochs, but also providing an opportunity for pruning model parameters. In the context of genetic algorithms, pruning is accomplished through viability or sexual selection.
This work uses NN implementations which load a model checkpoint file, train the model for one epoch, and then save the resultant model to a checkpoint file. The GA implementation can then manage population members' model parameters by managing checkpoint files. A shortcoming of PBT is that no easy and effective way to apply it to the optimization of NN topology has been demonstrated, as a change in topology creates a structural mismatch with previous checkpoints.
However, once a reasonable topology is found, PBT can fine-tune the HP schedule along with an optimization of model parameters, resulting in networks with accuracies much better than those attainable from traditional HPO alone.

\subsubsection{ResNet-20 on CIFAR-10} \label{sect:resnet20-cifar10}
\begin{figure}[!ht]
\centering
\includegraphics[width=1.0\textwidth]{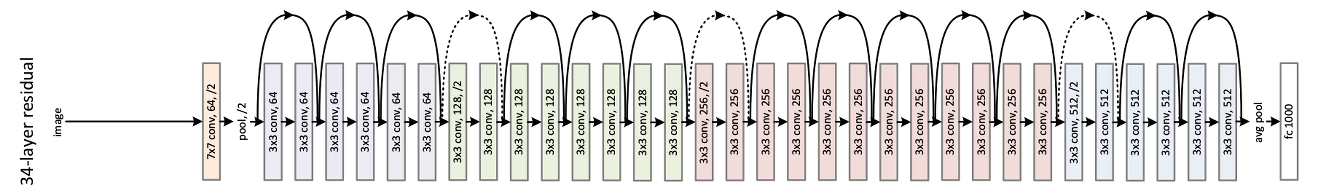}
\caption{\label{fig:resnet}ResNet neural network architecture (ResNet-34 pictured; image from \cite{he2016deep}).}
\end{figure}

\begin{figure}[!ht]
\centering
\begin{tabular}{ccccc}
\includegraphics[width=1cm]{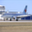} & 
\includegraphics[width=1cm]{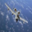} & 
\includegraphics[width=1cm]{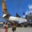} & 
\includegraphics[width=1cm]{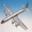} \\ 
\includegraphics[width=1cm]{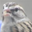} &
\includegraphics[width=1cm]{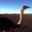} & 
\includegraphics[width=1cm]{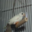} & 
\includegraphics[width=1cm]{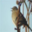} \\
\includegraphics[width=1cm]{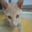} &
\includegraphics[width=1cm]{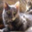} & 
\includegraphics[width=1cm]{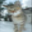} & 
\includegraphics[width=1cm]{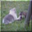} \\
\end{tabular}
\caption{\label{fig:cifar10}Selected example images from the CIFAR-10 data set (images from \cite{krizhevsky2009learning}).}
\end{figure}

The ResNet architectures were introduced as a solution to the difficulties of training deep neural networks \cite{he2016deep}.  ResNets accomplish this via skip-connections, illustrated in Figure \ref{fig:resnet}, which allow the networks to learn features centered around the identity rather than centered around zero.
This improves gradient transmission to the early layers and ensures that extraneous layers do not harm convergence. 
Our work investigates ResNet-20 with the CIFAR-10 dataset,  consisting of 60,000 RGB images divided into a training set of 50,000 and a testing set of 10,000 \cite{krizhevsky2009learning}.
These image are $32\times32$ pixels and fall into ten classes of objects, including airplanes, birds, and cats, as presented in Figure \ref{fig:cifar10}.

\begin{figure}[!ht]
\vspace{0.5\baselineskip}
\centering
\begin{tabular}{cc}
\includegraphics[width=0.475\textwidth]{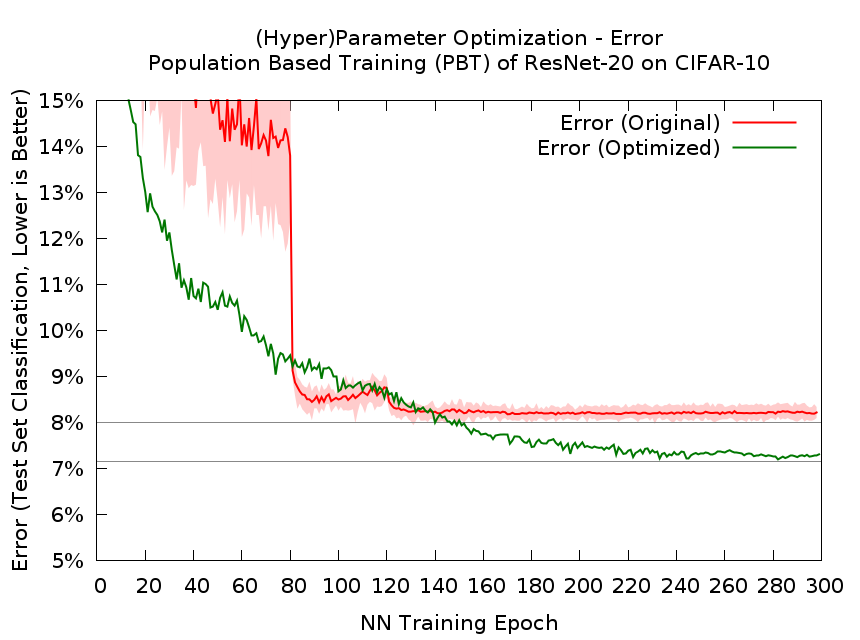} & \includegraphics[width=0.475\textwidth]{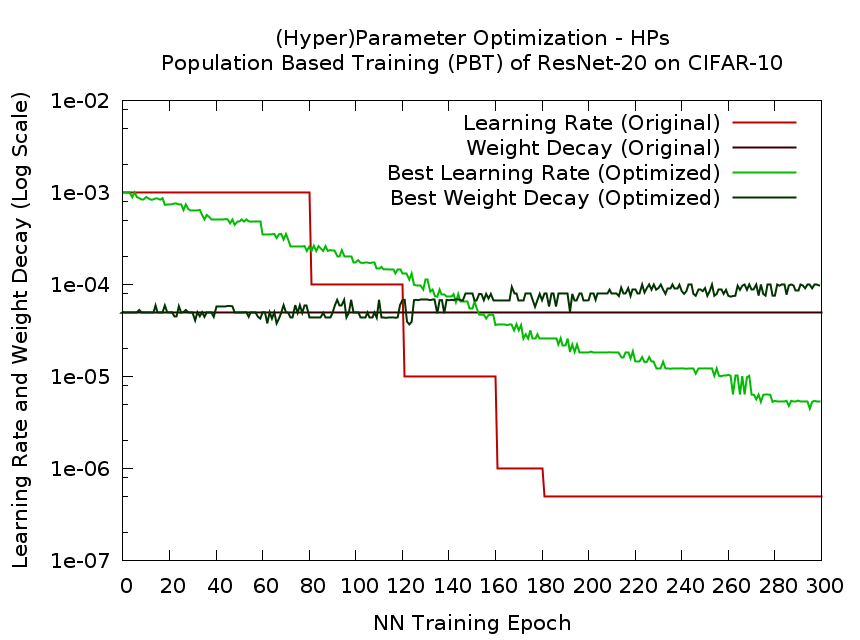} \\
\end{tabular}
\caption{\label{fig:resnet20_cifar10} Top-1 test set classification error across SGD training epochs for original HPs and evolved HPs for ResNet-20 on CIFAR-10 (left), and the corresponding hyperparameter schedule from PBT training (right).}
\end{figure}

In this example, our genetic algorithm is applied to ResNet-20 through the optimization of the training hyperparameters \textit{learning\_rate} and \textit{weight\_decay}, as well as the model parameters through the management of checkpoint files on \textit{Cicero} (see Section \ref{sect:exp_platform}). Figure \ref{fig:resnet20_cifar10} shows average top-1 classification error from ten different training runs on the left plot for the test set across SGD training epochs for both the original HPs (dark red) as well as evolved HPs (green).
Minimum and maximum error for the original HPs acquired from the training runs is depicted by the light red envelope in Figure \ref{fig:resnet20_cifar10}. These results show an optimized converged classification error of just over 7\% compared to 8\% for the original, representing a 12.5\% relative improvement. 

A comparison of the original and optimized training schedules is presented on the right of Figure \ref{fig:resnet20_cifar10}.  Our results show an exponential decay of \textit{learning\_rate} is better than the original step-wise decay, and a slow, nearly exponential growth of \textit{weight\_decay} is better than the original constant.
The original step-wise \textit{learning\_rate} decay and constant \textit{weight\_decay} represent accepted best practice, upon which our GA was able to improve. HPO approaches like PBT applied to already reasonable hyperparameter settings fine-tune them across NN training epochs for increased accuracy.
Indeed, the test set error for our optimized ResNet-20 on CIFAR-10 is better than the original paper's results for ResNet-56 on the same dataset \cite{he2016deep}.

\subsubsection{Capsule Net on CIFAR-10} \label{sect:capsnet_cifar10}
Capsule networks, introduced by Hinton at NIPS in 2017 \cite{capsnet_hinton}, are poorly characterized due to their recent development, making them an interesting candidate for further study with HPO. These networks accomplish improved feature detection over traditional convolution by dividing layers into groups of neurons called capsules.
Capsule layers are different from traditional convolutional networks which may be viewed as representing specific features in an example as activations of individual neurons. Instead, a capsule layer utilizes layer-wise activations over groups of neurons and vector normals of the group to capture both the likelihood of a feature being present in the example and the parameters which represent the specific feature. 

\begin{figure}[!ht]
\centering
\includegraphics[width=0.9\textwidth]{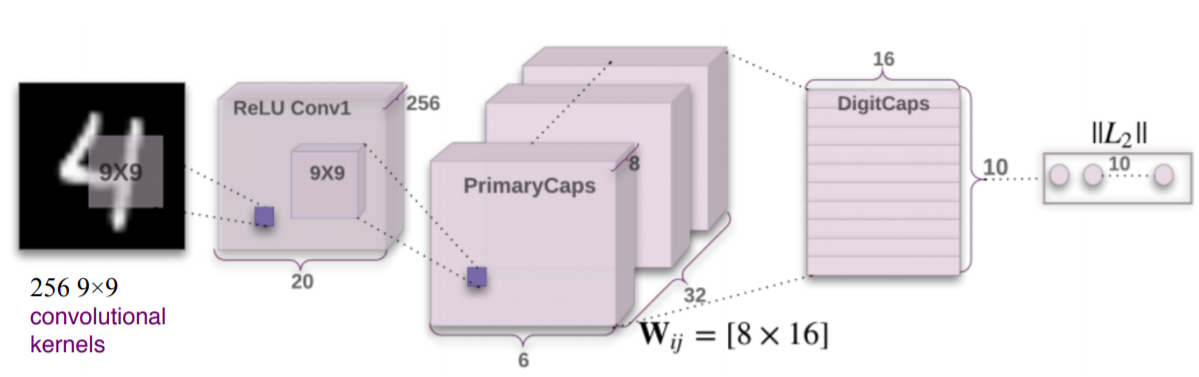}
\caption{\label{fig:capsnet_arch} Capsule network architecture similar to that used in this work (image from \cite{capsnet_hinton}).}
\end{figure}

Each capsule outputs a vector, and the length of each vector is used to determine the probability that a feature is present in a particular capsule. A squashing function is used to determine the vector output's length, which includes the free parameter $q$, as shown in Equation \ref{eqn:squash}, where  $v_{j}$ is vector output of capsule $j$, and $s_{j}$ is its total input. The free parameter $q$ is set to 1 in the original paper \cite{capsnet_hinton}, while the implementation used in this work sets $q$ to $\frac{1}{2}$ \cite{chollet2015keras}.

\begin{equation}\label{eqn:squash}
v_{j}=\frac{\vert{\vert{s_{j}}\vert{\vert}}^2}{q+\vert{\vert{s_{j}}}\vert{\vert}^2} \cdot \frac{s_{j}}{\vert{\vert{s_{j}}}\vert{\vert}}
\end{equation}

In this experiment, we apply HPO with PBT to $q$, \textit{learning\_rate}, \textit{weight\_decay}, and \textit{routings}. The \textit{routings} hyperparameter specifies the number of dynamic routings used to iteratively determine the coupling coefficients between capsules.
The network topology, a simplified version of which is depicted in Figure \ref{fig:capsnet_arch}, consists of a 4-layer convolutional network with ten capsules in the capsule layer, each with a dimension of 16.

\begin{figure}[!ht]
\centering
\begin{tabular}{cc}
\includegraphics[width=0.475\textwidth]{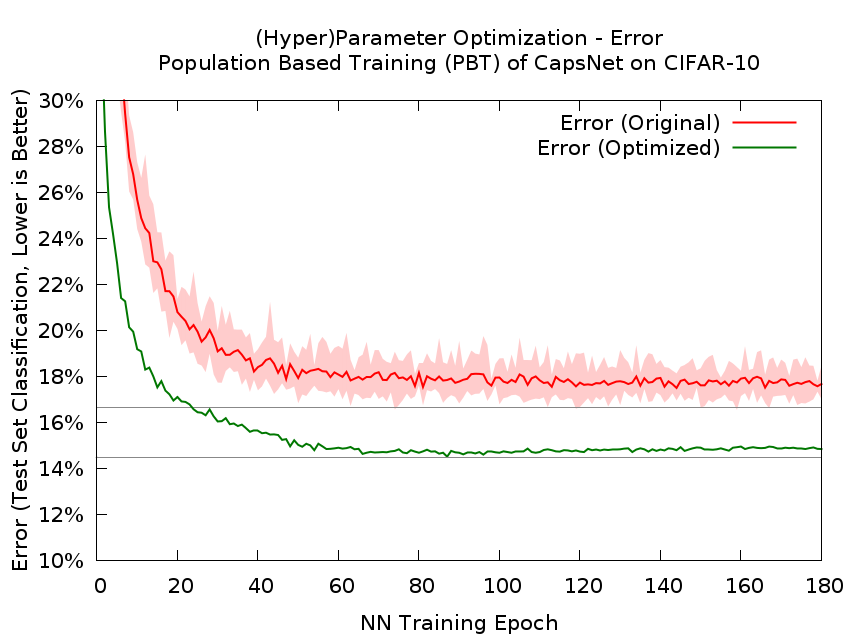} & \includegraphics[width=0.475\textwidth]{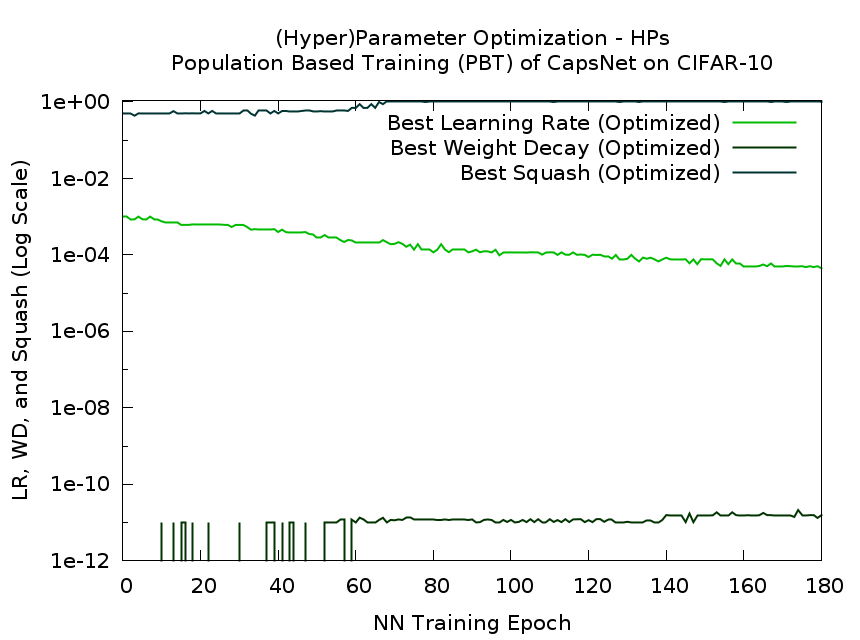} \\
\end{tabular}
\caption{\label{fig:capsnet_cifar10} Top-1 test set classification error across SGD training epochs for original HPs and evolved HPs for CapsuleNet on CIFAR-10 (left). Corresponding hyperparameter schedule from PBT training (right).}
\end{figure}

Figure \ref{fig:capsnet_cifar10} shows average top-1 classification error from ten different training runs on the left plot for the test set across SGD training epochs for both the original HPs (dark red) as well as evolved HPs (green).  Minimum and maximum error for the original HPs acquired from the training runs is depicted by the light red envelope.
These results from \textit{Cicero} (see Section \ref{sect:exp_platform}) show an optimized converged classification error of 14.5\% compared to 16.75\% for the original, representing a 15.5\% relative improvement. A training schedule was discovered for all included HPs, with some common features such as a decaying learning rate and a growing weight decay. Interestingly, the ``correct'' value of $q$, as initially proposed by Hinton, is rediscovered.

\subsubsection{NMT on Vi to En} \label{sect:nmt_vien}
In this section, a neural machine translation (NMT) network is trained to translate from Vietnamese to English \cite{luong17}. The example NN uses a sequence-to-sequence (Seq2Seq) recurrent neural network (RNN) architecture and is trained on a data set consisting of 2,821 sentences split into a training set of 1,553 and a test set of 1,268. These sentences are formed from a vocabulary of 7,709 Vietnamese words and 17,191 English words.

The optimization of an RNN with PBT highlights the need for efficient management of checkpoint data, as RNN model sizes are typically larger than those of convolutional neural networks. Without sufficient I/O bandwidth, checkpoint management can become a bottleneck for training performance. While the training data set is typically read-only, management of checkpoint files is read-write. Our work utilizes a shared Lustre parallel filesystem \cite{schwan2003lustre} on each of the test platforms for all checkpoint file management.

\begin{figure}[!ht]
\centering
\includegraphics[width=0.4\textwidth]{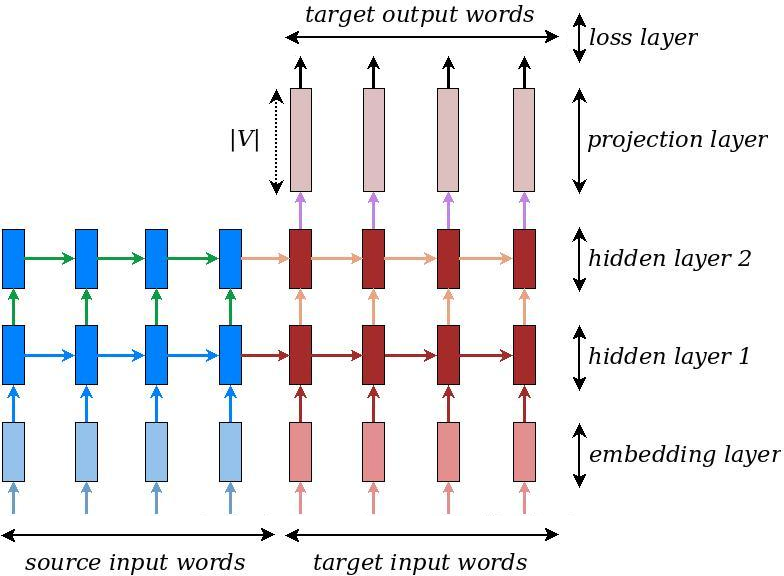}
\caption{\label{fig:nmt_arch}NMT Seq2Seq neural network architecture (image from \cite{luong17}).}
\end{figure}

\begin{figure}[!ht]
\centering
\begin{tabular}{cc}
\includegraphics[width=0.475\textwidth]{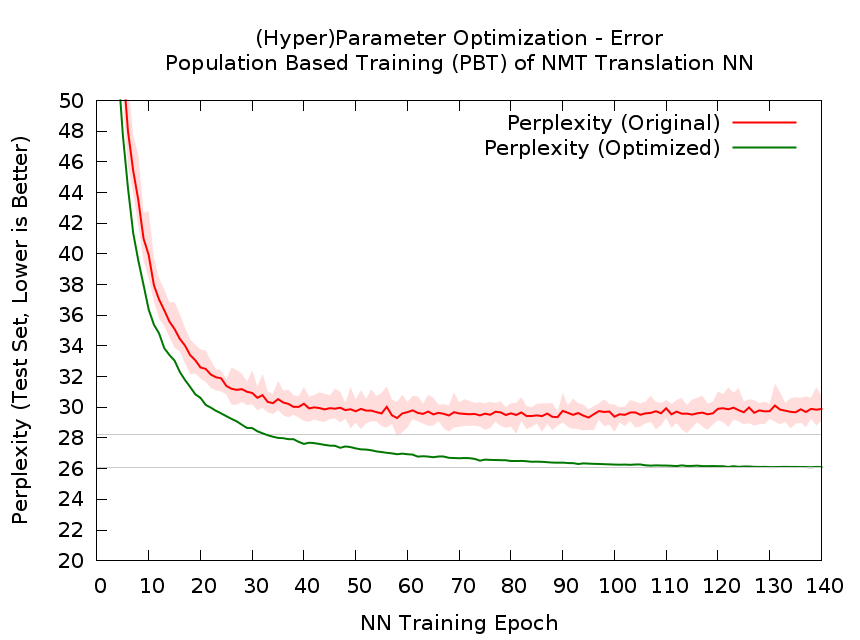} & \includegraphics[width=0.475\textwidth]{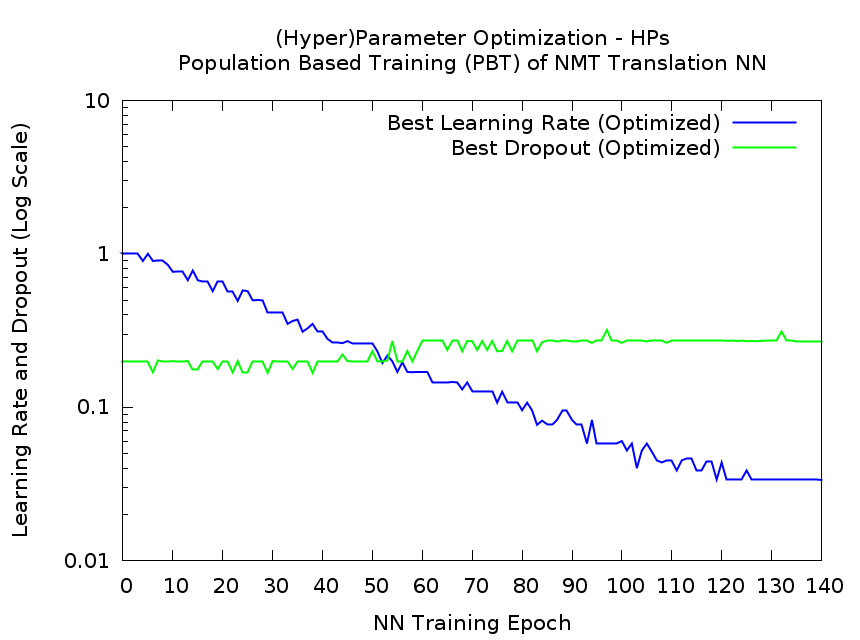} \\
\end{tabular}
\caption{\label{fig:nmt_pbt} PBT training of NMT showing both the perplexity on the test set during training epochs as well as the evolved hyperparameter training schedule.}
\end{figure}

Our genetic algorithm is applied to the NMT model parameters through checkpoint management on \textit{Cicero} (see Section \ref{sect:exp_platform}) as well as the \textit{learning\_rate} and \textit{dropout} schedules, with results presented in Figure \ref{fig:nmt_pbt}.
Similar to what is shown in Section \ref{sect:resnet20-cifar10}, the GA discovers an exponential decay of \textit{learning\_rate} is better than the original constant, but this time discovers a different decay rate.
Additionally, the GA is able to discover a novel HP schedule for \textit{dropout}, indicating that an increase partway through SGD training improves upon the original constant.

This new HP training schedule, along with an optimization of the model parameters, results in an evolved converged perplexity of just over 26.  Compared to the original perplexity of just over 28, this is an improvement of over 7\% (these results are conservative due to the inclusion of an outlier from the ten runs performed using original hyperparameters).

\section{Discussion}
Trends can be identified which hold across multiple neural network architectures as well as different data sets.
Both ``best practices'' and hyperparameter settings taken from papers are often suboptimal and can be significantly improved.
HPO with PBT frequently finds a training schedule involving an exponential decay of the learning rate across NN training epochs, although different networks appear to do best with different decay rates.
Experiments show an increasing weight decay schedule to be beneficial, though the precise shape of the curve appears to be network and data set specific.
Finally, the experiments summarized in Table \ref{tab:result_summary} demonstrate that the same core genetic algorithm can be successfully applied both to traditional HPO including topology HPs as well as to HPO with PBT.

\begin{table}[ht!]
\centering
\begin{small}
\begin{tabular}{llll|llr}
\textbf{Section} & \textbf{NN} & \textbf{Type} & \textbf{Dataset} & \textbf{FoM} & \textbf{HPO} & \textbf{Improvement} \\
\hline
$\vphantom{|^|}$\ref{sect:candle} & CANDLE P3B1 & dense & clinical reports & time to error & traditional & 78.2\% \\
\ref{sect:PRV_atlas} & RPV & CNN & ATLAS & final error & traditional & 80.7\% \\
\ref{sect:lenet_mnist} & LeNet & CNN & MNIST & time to error & traditional & 83.5\% \\
\ref{sect:resnet20-cifar10} & ResNet-20 & CNN & CIFAR-10 & final error & PBT & 12.5\% \\
\ref{sect:capsnet_cifar10} & CapsNet & capsule & CIFAR-10 & final error & PBT & 15.5\% \\
\ref{sect:nmt_vien} & NMT & RNN & Vi$\rightarrow$En & final error & PBT & 7.0\% \\
\end{tabular}
\end{small}
\vspace{.25\baselineskip}
\caption{\label{tab:result_summary}Improvement summary for experiments performed in Section \ref{sect:experiments} .}
\end{table}

\section{Future Work}

\subsection{Improved Recombination}
In the current genetic algorithm, two HP parents mate and recombine to create two genotypes, but only a single randomly chosen genotype is kept. Future versions of the genetic algorithm may benefit from the production of a child for each of the resultant genotypes. It is expected that keeping both children until after viability selection may help prevent beneficial genes from being lost from the population.
Additionally, there are two common types of crossover: one in which a crossover event encompasses a single locus, and one in which a crossover event means a switching of the source of all following loci until the next crossover event. The current GA applies each crossover event to a single locus, while the other approach may prove more effective.

\subsection{Demes}
Most if not all previous work applying genetic algorithms to hyperparameter optimization have used GAs employing only a single interacting population. In contrast, many biological models explicitly carve up geographic space into mostly-separate populations with some amount of migration between them.
These separate populations are referred to as \textit{demes}. While biologically realistic, the impetus for the inclusion of demes into a GA with an HPO focus is more about the problem of local optima.

\begin{figure}[!ht]
\centering
\includegraphics[width=0.15\textwidth]{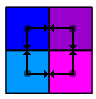}
\caption{\label{fig:migration}A possible migration strategy copies the best individuals to neighboring demes at regular generational intervals. The different colors indicate different dominant ``species'' in each of the local subpopulations.}
\end{figure}

A single population tends to cluster around a small number of candidate solutions, as less fit individuals are pruned from the population via selection.  Splitting the population into a number of weakly-interacting demes allows for demes to cluster around different local optima, encouraging better coverage of the HP search space.
One possible approach places demes on a Cartesian grid and allows migration between neighbors, depicted in Figure \ref{fig:migration}. Previous work suggests dispersal should not be so high as to prevent local adaptation due to gene flow \cite{birand2011patterns}. However, some gene flow is expected to be desirable as a means to prevent any one deme from becoming stuck in an unpromising area of the search space.

\subsection{Ploidy and Crossover of NN Parameters via Ensembles}
The term \textit{ploidy} refers to the number of sets of homologous chromosomes in an organism's nuclei.  As a familiar example, humans are diploid, having two complete sets of 23 chromosomes: one set from a mother and one set from a father.  However, many organisms found in nature are instead haploid, having only one set of chromosomes, or even polyploid with more than two.
Research suggests that diploid individuals are expected to have an advantage when evolutionary change is limited by mutation, whereas haploid individuals should have an advantage when evolutionary change is instead limited by selection \cite{otto2008evolution}. EvoDevo currently utilizes haploid individuals, and an exploration of ploidy may be warranted.

In isolation, exploring ploidy in our genetic algorithms is not expected to yield large gains, as there is some indication that haploid individuals are able to adapt faster in a number of situations \cite{otto2007evolutionary}. However, the concept of ploidy becomes interesting when combined with NN ensembles.  
That is, in the current GA, one can consider the individuals to be haploid with respect to their parameters, having an ensemble of size one, which always comes from a single parent. While there is sexual reproduction of individuals with respect to hyperparameters and with respect to the pairing of hyperparameters and parameters, there is no crossover inside the parameters themselves.

In large part this lack of parameter crossover is due to the lack of a good way to perform crossover between two different NNs, each with their own training history.  That is, swapping parts of one net into another is problematic, as for example each subnet may expect different input features and provide different output features which may conflict with the rest of the resultant NN.
Future work will explore individuals having an ensemble of two or more NNs, with some members coming from one parent and some from the other(s).  While this approach does not enable crossover \textit{within} an ensemble member, it does allow crossover among ensemble members, providing for sexual reproduction of not just hyperparameters, but parameters as well.

\section{Conclusions}
This work introduces an advanced genetic algorithm for hyperparameter optimization and applies it both to traditional HPO searches including topology as well as population-based training for the creation of training schedules. We demonstrate that optimization of (hyper)parameters can provide improvements in both final accuracy as well as time to fixed accuracy over default ``best practices'' for a wide range of deep neural network architectures, including convolutional neural networks, recurrent neural networks, dense neural networks, and capsule networks. Finally, we propose future research directions which are expected to provide additional gains in HPO efficacy.

\section{Acknowledgements}
This research used resources of the National Energy Research Scientific Computing Center (NERSC), a U.S. Department of Energy Office of Science User Facility operated under Contract No. DE-AC02-05CH11231 (see Section \ref{sect:exp_platform} for more details on \textit{Cori}).

\section{References}
\vspace{1\baselineskip}
\bibliographystyle{abbrv}
\begingroup
\renewcommand{\section}[2]{}%
\begin{footnotesize}
\bibliography{bibliography}
\end{footnotesize}
\endgroup

\end{document}